\documentclass{ecai} 

\usepackage{latexsym}
\usepackage{amssymb}
\usepackage{amsmath}
\usepackage{amsthm}
\usepackage{booktabs}
\usepackage{enumitem}
\usepackage{graphicx}
\usepackage{color}

\usepackage{xcolor}
\definecolor{myblue}{RGB}{143, 170, 220}
\definecolor{mygreen}{RGB}{169, 209, 142}

\usepackage{CJKutf8}
\usepackage{colortbl}
\usepackage{multirow}

\usepackage{pgfplots}

\usepackage{makecell}

\newcommand{\BibTeX}{B\kern-.05em{\sc i\kern-.025em b}\kern-.08em\TeX}

\begin{document}


\begin{frontmatter}




\title{A Coin Has Two Sides: A Novel Detector-Corrector Framework for Chinese Spelling Correction}



\author[A]{\fnms{Xiangke}~\snm{Zeng}}
\author[A]{\fnms{Zuchao}~\snm{Li}\thanks{Corresponding Authors. Email: zcli-charlie@whu.edu.cn. This work was supported by the National Natural Science Foundation of China (No. 62306216, No. 72074171, No. 72374161), the Natural Science Foundation of Hubei Province of China (No. 2023AFB816), the Fundamental Research Funds for the Central Universities (No. 2042023kf0133).}}
\author[A]{\fnms{Lefei}~\snm{Zhang}\footnotemark[*]}
\author[B]{\fnms{Ping}~\snm{Wang}} 
\author[C]{\fnms{Hongqiu}~\snm{Wu}} 
\author[C]{\fnms{Hai}~\snm{Zhao}} 

\address[A]{School of Computer Science, Wuhan University}
\address[B]{School of Information Management, Wuhan University}
\address[C]{Department of Computer Science and Engineering, Shanghai Jiao Tong University}


\begin{abstract}
Chinese Spelling Correction (CSC) stands as a foundational Natural Language Processing (NLP) task, which primarily focuses on the correction of erroneous characters in Chinese texts. Certain existing methodologies opt to disentangle the error correction process, employing an additional error detector to pinpoint error positions. However, owing to the inherent performance limitations of error detector, precision and recall are like two sides of the coin which can not be both facing up simultaneously. Furthermore, it is also worth investigating how the error position information can be judiciously applied to assist the error correction.
In this paper, we introduce a novel approach based on error detector-corrector framework. Our detector is designed to yield two error detection results, each characterized by high precision and recall. Given that the occurrence of errors is context-dependent and detection outcomes may be less precise, we incorporate the error detection results into the CSC task using an innovative feature fusion strategy and a selective masking strategy. Empirical experiments conducted on mainstream CSC datasets substantiate the efficacy of our proposed method.
\end{abstract}

\end{frontmatter}


\section{Introduction}

\begin{CJK*}{UTF8}{gbsn}

Chinese Spelling Correction (CSC) aims to detect and correct erroneous characters in given Chinese sentences, which is a fundamental NLP task and plays an indispensable role in many NLP downstream tasks ~\cite{martins2004seq,gao2010large}, also a part of Grammer Error Correction (GEC) ~\cite{lizuchaoGEC}.

Up to now, CSC is still a challenging task. Chinese originated from pictograms, so the shapes and sounds of the characters are closely related to their meanings. Chinese always consists of many consecutive characters without separators ~\cite{wang-etal-2023-enhancing,huang2023frustratinglyeasyplugandplaydetectionandreasoning}, which makes the CSC method must be able to discern errors based on information in the context, rather than directly finding spelling errors from independent words.

In recent years, pre-trained language models (PLMs) have been quickly developed ~\cite{transformers,bert}. Current state-of-the-art methods regard CSC as a sequence tagging task and fine-tune BERT-based models on sentence pairs like Machine Translation (MT) task ~\cite{lizuchaoMT,guo-etal-2021-global,spellbert,li-2022-uchecker}. The source sentence is directly fed into the model and the target sentence is output. Some other works use an error detector as the preliminary for correction which turns the CSC into a two-stage pipeline~\cite{mdcspell,wei-etal-2023-ptcspell}. The detection results produced by the detector indicate which characters are incorrect. However, using detector does not consistently lead to substantial performance improvements, which could be attributed to two potential reasons.

\begin{figure}[t]
\centering
\includegraphics[width=0.95\linewidth]{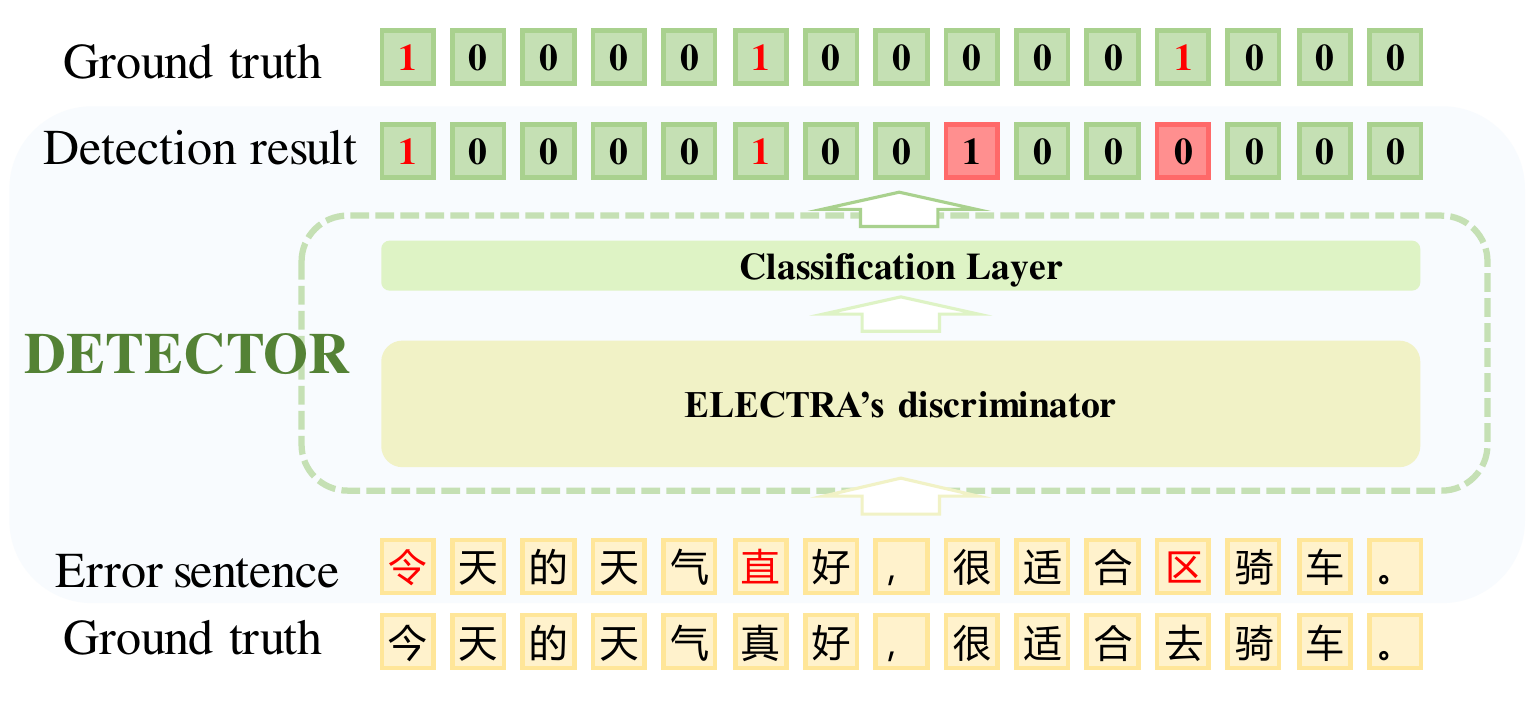}
\caption{An example of detector in CSC task. The output of detector is a series of zeros and ones, in which zero means the character in corresponding position is right while one means error. In the source sentence, ``令", ``直" and ``区" are wrong characters, but the detector fails to detect ``区" and judges the right character ``很" to be wrong character by mistake. }
\label{fig1}
\vspace{2em}
\end{figure}

The primary reason is that the detector may not be strong enough, and a poor detector even has a negative impact on the subsequent error correction process. As shown in Figure \ref{fig1}, while using ELECTRA as a detector for error detection, it always fails to work as expected. In some preliminary experiments, it is observed that an error detection network may yield poorer results compared to the direct judgment made by the correction network regarding whether a character has been altered. Consequently, such error detection results demonstrate limited or even negative enhancement in error correction capabilities.

Otherwise, the use of detectors in the existing works is not satisfactory. Existing methods basically use the detector at the character level, that is, directly indicate in some way where the error occurred in the original sentence, which can make the error information attached to the corresponding position to assist correction. We find that in the error correction of a Chinese sentence, there is a strong relationship between the error character and the context, and only applying the indication to the corresponding position cannot indicate the model to focus on this.

In order to solve these problems, we propose a novel detector-corrector-based method that further improves the effectiveness of the detector and optimizes the utilization of detection results. Specifically, we constrain the detector to obtain two sets of detection results with high precision and high recall. For the high-precision detection results obtained, we apply the Fuzzy Indication technique on them and perform feature fusion with the original sentences. For the high-recall detection results obtained, we mask the corresponding positions and their context within a sentence, then concatenate this sentence after the original sentence.

The principal contributions of this paper can be summarized as follows:

1. We design the detector to generate high-precision and high-recall detection results, to some extent addressing the trade-off between precision and recall in the detector.

2. We consider the contextual relevance of errors and the inherent detection inaccuracies, and subsequently design two strategies for the utilization of detection results. This approach renders the error correction process more rational and adaptive.

3.  We investigate the performance of our method both quantitatively and qualitatively. The experimental results show the superiority of our method on mainstream benchmarks.

\end{CJK*}

\section{Related Work}

Chinese spelling correction is a challenging natural language processing task, which plays a very important role in many downstream tasks. It needs to detect and correct errors in a given Chinese sentence. Most of the early works use unsupervised language models and rules for detection and correction, and use the perplexity of language model for determination ~\cite{yeh-etal-2013-chinese,yu-li-2014-chinese,xie-etal-2015-chinese}. 

Recent years, pretrained language models such as BERT~\cite{bert} have shown strong performance on NLP-related tasks. Many works use BERT-like models in CSC to directly correct the whole sentence to be error-free. Some methods introduce phonemes and glyphs as additional information to improve CSC. SpellGCN~\cite{spellgcn} incorporates phonological and visual similarity knowledge into BERT via a specialized graph convolutional network. PLOME~\cite{plome} uses a GRU network to obtain the phonological and visual embeddings, and then performs feature fusion in the network training process. SCOPE~\cite{scope} builds two parallel decoders to predict target characters and pinyin respectively and balances them adaptively according to the phonetic similarity between input and target characters. DORM~\cite{liang-etal-2023-disentangled} introduces a pinyin-to-character objective to ask the model to predict the correct characters based solely on phonetic information, then designs a self-distillation module to ensure that semantic information plays a major role in the prediction.

Some works turn the CSC into a two-stage pipeline by using an error detector as a preliminary task and then use the detection results to assist the subsequent error correction process. Soft-Masked BERT~\cite{softmaskbert} masks the detected wrong characters from error detection with error probability and then turns the masked input into the BERT model for error correction. DCSpell~\cite{10.1145/3404835.3463050} masks the wrong characters detected from the detector in source sentence and then concatenate it with raw source sentence as inputs of corrector. MDCspell~\cite{mdcspell} uses BERT to capture the visual and phonological features of the characters and integrated the hidden states of the detector to reduce the impact of error. PTCSpell~\cite{wei-etal-2023-ptcspell} designs two pre-training objectives to capture pronunciation and shape information in Chinese characters for CSC, and uses error detection results to process the error correction results. 

While the utilization of detectors in the aforementioned methods has indeed led to improvements in CSC task performance, they still have shortcomings. These drawbacks include the potential for erroneous error detection results to misguide the process. Additionally, the effective application of error detection information warrants further investigation.
To address these issues, we present a novel approach based on the detection-correction structure. This method capitalizes on error detection to a greater extent, thereby enhancing the overall effectiveness of error correction through the detection process.

\begin{figure*}[t]
\centering
\includegraphics[width=0.85\linewidth]{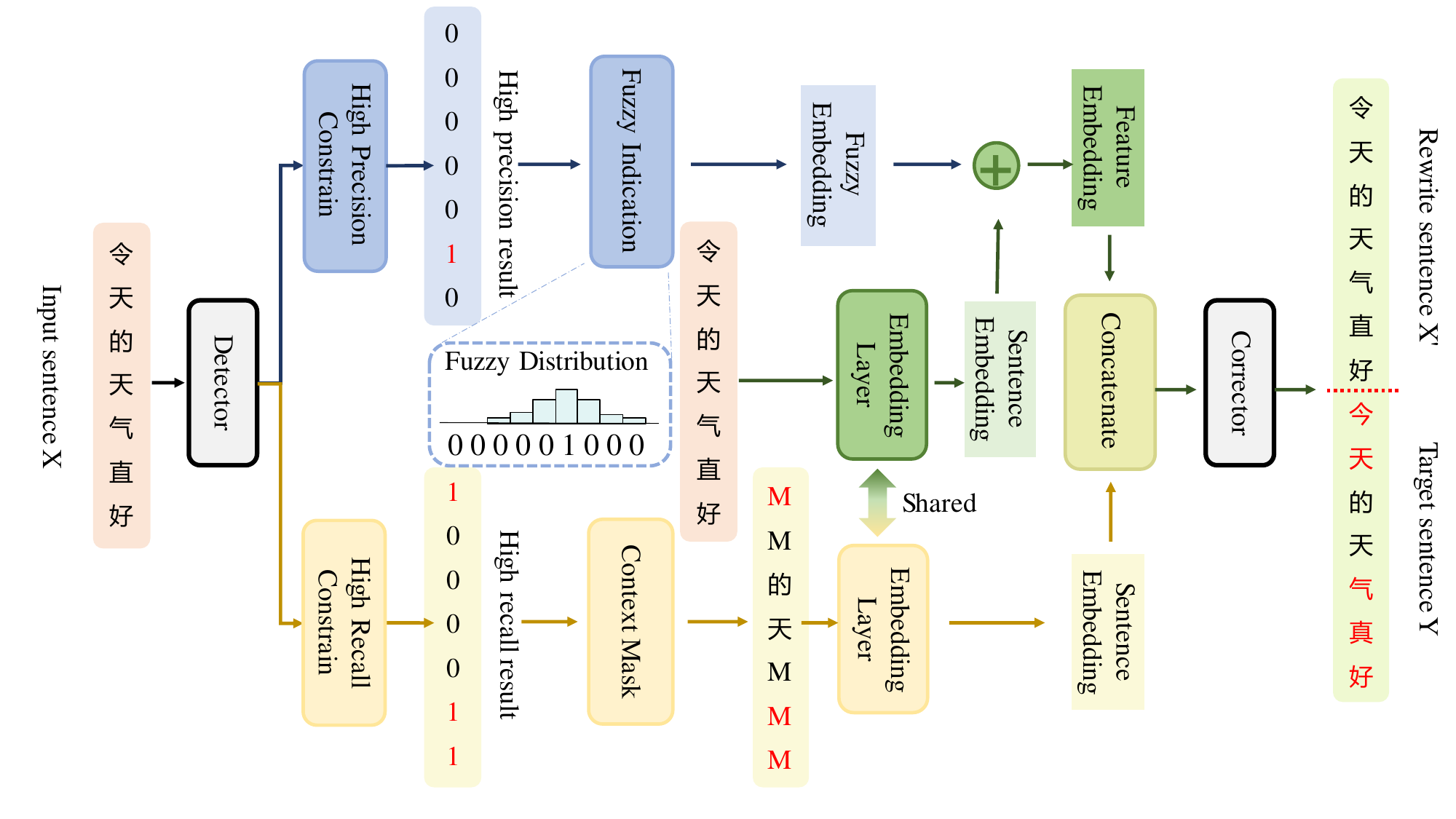}
\caption{The architecture of COIN. After getting high precision and high recall results from Detector, we separately use EP and SM. The EP strategy is used to "point out the errors" which informs the corrector to focus on wrong characters, and the SM strategy is used to "set blanks" for the corrector to "fill in". The latter half of the output sentence represents the correction result. Characters in red in the final output indicate tokens generated from masks by the Corrector. }
\label{fig2}
\vspace{2em}
\end{figure*}

\section{Methodology}
\label{sec:methodology}

\subsection{Problem Definition}
\label{sec:problem_definition}
Chinese spelling correction is a task to detect and correct erroneous characters in given Chinese sentences. Given a Chinese sentence $X=\{x_1,x_2,..,x_n\}$ of $n$ characters that may include erroneous characters, we use $Y=\{y_1,y_2,..,y_n\}$ to represent the corresponding correct sentence. The objective of CSC is to detect and correct the erroneous characters by generating a prediction $\hat{Y}=\{\hat{y}_1,\hat{y}_2,..,\hat{y}_n\}$ for the input $X$, where $\hat{y}_i$ is the character predicted for $x_i$.  

\subsection{Motivation and Preliminary Experiments}
\label{sec:motivation}

Given the detection results, we design two strategies to leverage error detection results for enhancing the error correction process. 
The first strategy is the Error Position Information Fusion Strategy (EP), which entails feature fusion to identify the presence of an error by adjusting the embedding of the error token.
The second strategy is the Selective Masking Strategy (SM), which involves masking tokens corresponding to error positions, thereby indicating the positions of errors and guiding the model in correction process.

Methods based on detection-correction structure highly dependent on the performance of the error detector. Existing methods always set detector as simplistic networks lacking additional design, which results in suboptimal detection capabilities. 

The evaluation of an error detector hinges on two critical metrics: precision and recall. As the old saying goes, "a coin has two sides". Precision and recall are like the two sides of the coin. In cases where achieving a better detector seems impossible, improving precision invariably involves being more selective and rejecting ambiguous predictions, which subsequently leads to a reduction in recall. Conversely, when precision increases, recall tends to decrease. There is no coin with both sides facing up, also no a method can improve both precision and recall of a detector while working within certain performance limitations. 

To explore the influence of detection result on error correction, we deliberately introduce detection errors to examine the attributes of the two strategies as preliminary experiments. 



 As shown in Table \ref{table_pre1}, when an error failed to be detected, indicating a correct character as an error will introduce significant precision degradation. 
This can be explained by the fact that when a correct token is mistakenly indicated as erroneous, it becomes challenging for the neural network to establish a reliable association between the indications and actual errors. 


\begin{table}[t]
\caption{Preliminary experiments of Error Position Information Fusion Strategy (EP) on ECSpell Law dateset. "Cor" represents the percentage of accurate detection results, "Wr" signifies the relative proportion of errors in the detection results, FP signifies False Positive, that is correct character which is detected as error. }
\vspace{2em}
\centering

\resizebox{0.81\linewidth}{!}{
\begin{tabular}{l|c|ccc|ccc}
\toprule
\multicolumn{2}{c}{EP}&\multicolumn{3}{c}{No FP}&\multicolumn{3}{c}{With FP}\\
\hline

Cor.&100&90&80&70&90&80&70\\
\hline
Wr.&0&0&0&0&10&20&30\\
\hline
F1&98.1&96.1&92.6&91.4&95.4&91.7&89.2\\
\bottomrule

\end{tabular}
}

\label{table_pre1}
\end{table}

\begin{table}[t]
\caption{Preliminary experiments of Selective Masking Strategy (SM) on ECSpell Law dateset. "Cor", "Wr" and "FP" are the same as Table \ref{table_pre1}. }
\vspace{2em}
\centering
\resizebox{0.81\linewidth}{!}{
\begin{tabular}{l|c|ccc|ccc}
\toprule
\multicolumn{2}{c}{SM}&\multicolumn{3}{c}{No FP}&\multicolumn{3}{c}{With FP}\\
\hline

Cor.&100&90&80&70&90&80&70\\
\hline
Wr.&0&0&0&0&10&20&30\\
\hline
F1&95.5&91.7&89.3&86.9&91.6&88.8&86.4\\
\bottomrule

\end{tabular}
}

\label{table_pre2}
\end{table}

 As shown in Table \ref{table_pre2}, the effectiveness of error correction is significantly impacted by unmasked errors, while correct character mistakenly masked doesn't seem to matter much. This can be succinctly understood as that model can easily generate the raw characters from mistakenly masked correct characters. But if a substantial number of incorrect words remain unmasked, the neural network is still compelled to decide whether to modify them when processing unmasked tokens. This sets an additional obstacle for the model. 

Hence, EP strategy anticipates that the detected error is indeed an error, while SM strategy anticipates that the unmasked character is correct. In the fact of that using the detection results can indeed benefit the error correction process, our SM strategy is more tolerant to worse precision of the detection, while the EP strategy is more tolerant to worse recall of the detection. The characteristics of these two strategies aligning with the goals of achieving high precision and high recall results, which can be obtained by our design of the detector. Therefore, we propose our method named COIN, reflecting this duality in error indication strategies. 

\subsection{Architecture}
\label{sec:model_detail}
As shown in Figure \ref{fig2}, our model is constructed with two parts: Detector and Corrector. We first use Detector to get our detection results, and then inject information of detection into Corrector by EP and SM. The combination of EP and SM can be expressed as a "point out the errors and fill in the blanks" game. 

\subsubsection{Detector}

Our detection network is based on the discriminator in ELECTRA~\cite{electra}. We use the pre-trained Chinese ELECTRA to initialize the weights of the detector. 
As in Figure \ref{fig1}, error detection is defined as a character-level binary classification task. Input sentences $X = \{x_1, x_2, ..., x_n\}$ are expressed as char-level tokens. Then the discriminator of ELECTRA encodes them as $H^c$. The classification layer can be presented as below:
\begin{equation}
    H^{'} = \textit{LayerNorm}(\textit{GELU}(W^{'} H^c + b^{'}))
\end{equation}

\begin{equation}
    H_{out} = W_{out} H^{'} + b_{out}
\end{equation}

\noindent where $c$ is the size of the hidden state from ELECTRA's discriminator, $\textit{GELU}$~\cite{gelu} is activation function and $LayerNorm$ means layer normalization~\cite{layernorm}. $H_{out} = \{h_1, h_2, ..., h_n\}$ is the classification layer's output representation of each character. After that, the probability of each character to be wrong is: 

\begin{equation}
    P^c (d_i = 1|X) = \textit{sigmoid}(h_i)
\end{equation}

\noindent where $d_i$ means the i-th of detection result and $P^d (d_i = 1|X)$ is a conditional probability indicating that $x_i$ is an wrong character. Then we can get two detection results by thresholds: 
\begin{equation}
	D(P^c) = \begin{cases}
	1, &\text{if\space} P^c > \lambda\\
	0, &\text{if\space} P^c < \lambda
		   \end{cases}
\end{equation}
\noindent where $\lambda \in \left\{ p, r\right\}$, $p$ and $r$ are two thresholds to constrain the detection results of detector. The higher threshold $p$ is applied such that predictions with high confidence are retained while others are ignored. When the lower threshold $r$ is applied, all predictions with certain degree of confidence are retained. The higher threshold can obtain higher precision at the expense of certain recall, and the lower threshold can obtain higher recall at the expense of certain precision.

\subsubsection{Corrector}
Our correction network is based on BERT~\cite{bert}. We use the pre-trained Chinese BERT to initialize the weights of the corrector. 
Given the high precision results and the high recall results mentioned above, we separately adopt our EP strategy and our SM strategy in error correction process.

For high precision results, we integrate the error detection results into the source sentence embeddings. Each error detection result shares the same length as its corresponding input sentence, with positions flagged as errors set to 1 and others set to 0. By directly adding these results to the source sentence embeddings in every channel dimension, we only alter the embeddings of tokens identified as errors. Consequently, the network can learn an error indication from these consistent offsets. Additionally, this operation causes the embedding of error tokens to diverge from the original semantics, making them more discernible and easier to be detected and corrected by the correction network.

\begin{CJK*}{UTF8}{gbsn}

\begin{table}[t]
\caption{An example of mistaken detection in SIGHAN14. The error occur in Chinese phrase ``山上" which is incorrectly written as ``山山". The detection network detects the first ``山" to be error because this phrase is possible to be a name of mountain. }
\vspace{2em}
\centering
\resizebox{0.7\linewidth}{!}{
\centering
\begin{tabular}{ll}
\toprule
Source  &我们马上去山\textcolor{red}{山}看风景。 \\
& We'll go to the mountains \textcolor{red}{mountain}\\ & to see the scenery.\\
\midrule
Target  &我们马上去山\textcolor{blue}{上}看风景。 \\
&We'll go to the mountains to see the \\ & scenery. \\
\midrule
Detection &0, 0, 0, 0, 0, 1, \textcolor{red}{0}, 0, 0, 0, 0 \\
\bottomrule

\end{tabular}
}

\label{table2}
\end{table}
\end{CJK*}

However, errors in Chinese text are inherently context-dependent. Simply applying information fusion exclusively affects characters identified as errors, without explicitly guiding the model to consider the surrounding context. Additionally, Chinese sentences are often composed of phrases, as illustrated in Table \ref{table2}, which may lead to deviation of the error position when an error occurs in a phrase. Consequently, even with a highly capable detector, there remains a discrepancy between actual error positions and detection results. Direct information fusion does not cope well with this situation. 

As shown in ~\cite{fsuie}, the traditional Dirac delta distribution may not perform optimally when dealing with the uncertainty or ambiguity associated with a single point indication. 
Also inspired by ~\cite{Li2020Data-dependent}, we introduce Fuzzy Indication (FI) strategy. We apply a mapping function, denoted as $F$, which map Dirac delta distribution to Gaussian distribution for error position prediction: 

\begin{equation}
	F(x) = \begin{cases}
	\epsilon, &\text{if\space} \epsilon > \theta\\
	0, &\text{if\space} \epsilon < \theta
		   \end{cases}
\end{equation}

\noindent where $\epsilon$ is corresponding Gaussian distribution variable of each character, and $\theta$ is sampling threshold which acts as a filter to eliminate information from unimportant positions. To make the continuous Gaussian distribution $N(\mu , \delta^2)$ adapt to our discrete case, given pre-determined variance $\delta$ and mean $\mu$ which is set to the real position of error, we set a sampling step $s$. Then $\epsilon$ of $x_i$ can be calculate as:
\begin{equation}
    \epsilon = Gauss(\mu + (i-g)s)
\end{equation}

\noindent where $g$ is the real error position, $Gauss$ is the probability density function. When there are two or more error in a sentence, the $\epsilon$ is calculated as the sum of values based on each. 
Subsequently, we obtain the Fuzzy Embedding $G(x)=[F(d_1),F(d_2),...,F(d_n)]$, which will be added to each channel of the sentence embedding to indicate the error position.

When the error position detection results deviate, the utilization of a Gaussian distribution aids in encompassing the potentially incorrect position within the error indication interval. This approach helps mitigate the impact of inaccurate indications. Empirical experiments demonstrate that our method can still yield favorable results even when there is a certain degree of deviation in the error position indications.

\begin{figure}[t]
\centering
\includegraphics[width=0.8\linewidth]{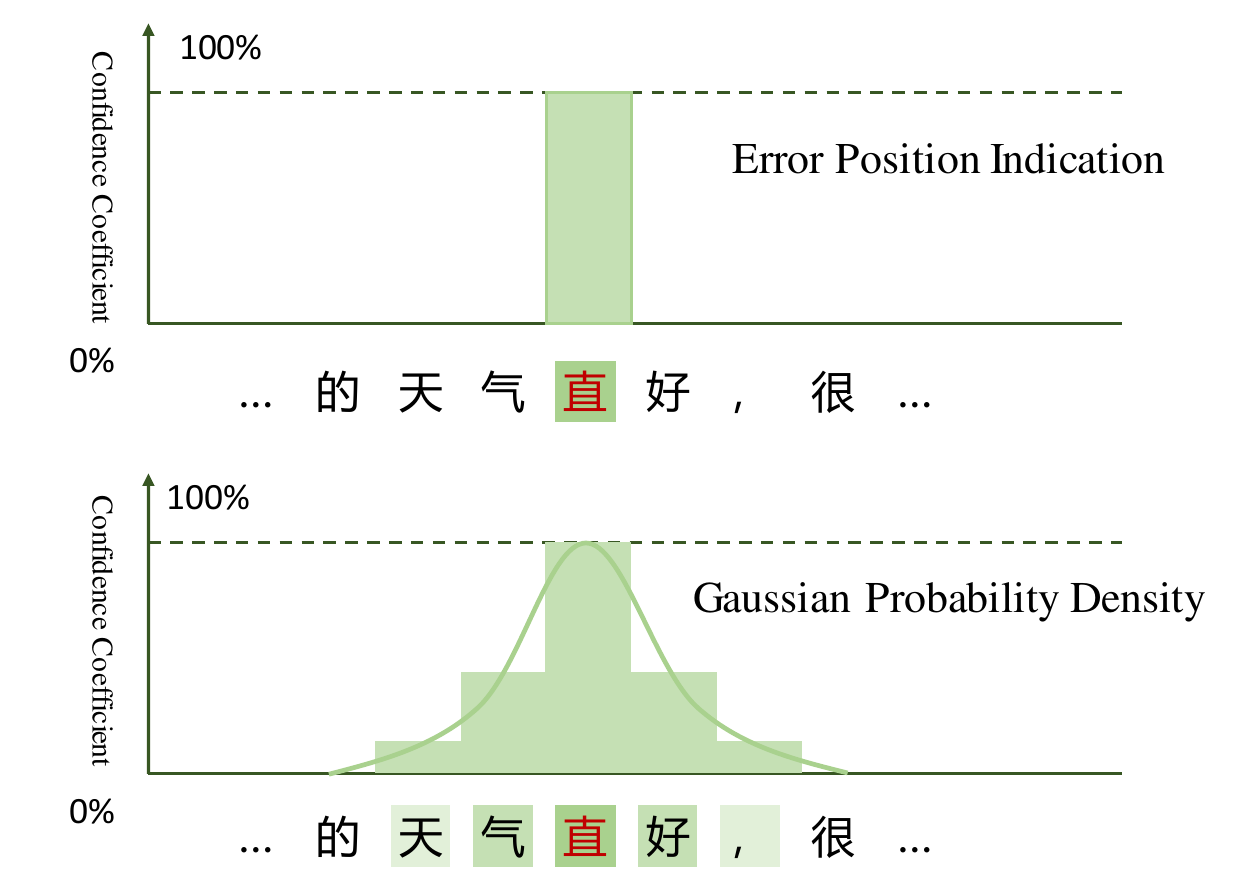}
\caption{Illustration of the mapping function to get Fuzzy Indication embedding of error position. }
\vspace{2em}
\label{fig3}
\end{figure}

For high recall results, we mask the positions corresponding to the error detection results. 
However, directly masking out the suspected error characters in the sentence will directly lose this part of the information. The occurrence of Chinese spelling errors is highly correlated with the similarity of correct and incorrect characters, thus we aim to retain the completeness of the entire sentence to leverage its full semantic context, while applying mask strategy to benefit from high recall results. 
Inspired by ReLM~\cite{relm} which transfers the original sentence to the semantic space and using the model to generate a semantically corrected sentence, we concatenate two source sentence and only adopt mask on the later. Specifically, we selectively mask the original sentence to get a partly masked sentence, and then concatenate it with the unaltered sentence. This is like rewriting the sentence once at the end of the original sentence, but leaving blank space for less certain places. This process can be expressed as follows:
\begin{equation}
    X_{SM} = \textit{Concat}(X, X_m)
\end{equation}

\noindent where $X_m$ is the sentence after selectively masking; $Concat$ means concatenation of sentences, $X_{SM}$ means the final output of selective masking strategy.
We replace the predicted error position tokens with masks, which aligns with the pre-training task of BERT and allows us to better leverage its capabilities.

Considering the potential deviations in detection results, as discussed previously, masking a single potentially erroneous character could severely mislead the error correction process, particularly when deviations occur. Also, the reliance on context is crucial since errors may depend on it, thus it is advisable to orient the corrector towards considering contextual information. 

Since the input contains the full original sentence, our method can easily generate characters without errors in source sentences from masks. In other words, the cost of masking additional potential errors is relatively low. Consequently, we can enhance the error tolerance to detection of our approach by extending the mask length. Specifically, characters situated within a specified proximity to the detected error are replaced with masks. This strategy not only prompts the model to consider the contextual backdrop of errors during prediction, as they are uniformly masked, but also exhibits resilience against inaccuracies in position prediction when detection is off the mark. This approach ensures that the model remains robust and efficient, even when faced with less precise error position. 

\section{Experiments}
\subsection{Datasets}
\textbf{ECSpell} ~\cite{ecspell} is a CSC benchmark with three domains, law (LAW, 1,960 training and 500 test samples), medical treatment (MED, 3,000 training and 500 test samples), and official document writing (ODW, 1,728 training and 500 test samples).

\noindent \textbf{SIGHAN} is a traditional CSC benchmark collected from the Chinese essays written by foreign speakers including SIGHAN13 ~\cite{sighan13}, SIGHAN14 ~\cite{sighan14}, and SIGHAN15 ~\cite{sighan15}. Following previous work ~\cite{pointer-networks,spellgcn,realise}, we merge the three SIGHAN training sets and another 271K pseudo samples generated by ASR or OCR ~\cite{wang271k} as training set. We evaluate our model on the test set of SIGHAN15. Since the original SIGHAN datasets are in Traditional Chinese, they are converted to Simplified Chinese by OpenCC. 

Following previous methods, we use the metrics of sentence-level precision, recall, and F1 to evaluate our model for correction.

\begin{table}[t]
\small
\caption{Overall results of baselines and BERT based COIN on ECSpell in precision, recall, and F1. The best results are shown in bold. The results of baselines are cited from the corresponding papers.}
\vspace{2em}
    \centering
    \resizebox{0.80\linewidth}{!}{
    \begin{tabular}{c | l | p{1cm}<{\centering} p{1.0cm}<{\centering} p{0.9cm}<{\centering} }
    \toprule
    {Dataset} & {Method} &    Prec. & Rec.  & F1 \\
     \midrule
     \multirow{7}{*}{LAW} & BERT      & 73.2 & 79.2 & 76.1 \\
                               & REALISE            & 63.1 & 61.6 & 62.3  \\
                               & MDCSpell  & 77.5 & 83.9 & 80.6 \\
                      
                      & ECSpell  & 78.3 & 74.9 & 76.6 \\
                               & RSpell    & 85.3 & 81.6 & 83.4 \\
                           & ReLM  & 89.9 & 94.5 & 92.2 \\
     \cline{2-5}
                               & COIN (ours)                 & \textbf{93.5} & \textbf{96.1} & \textbf{94.8}   \\
     \midrule
     \multirow{7}{*}{MED} & BERT      & 57.9 & 58.1 & 58.0  \\
                               & REALISE            & 55.0 & 46.0 & 50.1  \\
                               & MDCSpell  & 69.9 & 69.3 & 69.6 \\
                      & ECSpell  & 75.9 & 71.2 & 73.5  \\
                               & RSpell    & 86.1 & 77.0 & 81.3  \\
                           & ReLM  & 85.5 & 85.3 & 85.4  \\
     \cline{2-5}
                               & COIN (ours)                 & \textbf{92.2} & \textbf{94.0} & \textbf{93.1}    \\
     \midrule
     \multirow{7}{*}{ODW} & BERT      & 59.7 & 58.8 & 59.2  \\
                               & REALISE            & 55.0 & 50.6 & 52.7 \\
                               & MDCSpell  & 65.7 & 68.2 & 66.9  \\
                      & ECSpell  & 82.3 & 74.5 & 78.2 \\
                               & RSpell   & 89.0 & 79.9 & 84.2 \\
                           & ReLM  & 85.7 & 87.8 & 86.7 \\
     \cline{2-5}
                               & COIN (ours)                 & \textbf{91.8} & \textbf{91.5} & \textbf{91.7}      \\

    \bottomrule
    
    \end{tabular}
    }
    \label{table3}
    \vspace{0mm}
    
\end{table}

\begin{table*}[h]
\small
    \caption{Overall results of COIN and baselines on SIGHAN15 in precision, recall, and F1. The best results are shown in bold and the second-best results are underlined. The results of baselines are cited from the corresponding papers. "*" denotes that we post the raw results without post-processing for fairness. Following previous methods, we post both detection level results and correction level results. }
    \vspace{2em}
    \centering
    \resizebox{0.7\linewidth}{!}{
    \begin{tabular}{c | l | p{1cm}<{\centering} p{1.0cm}<{\centering} p{0.9cm}<{\centering} | p{1cm}<{\centering} p{1.0cm}<{\centering} p{0.9cm}<{\centering}}
    \toprule
    \multirow{2}{*}{Dataset} & \multirow{2}{*}{Method} & \multicolumn{3}{c}{Detection Level}  &  \multicolumn{3}{c}{Correction Level}  \\ 
     \cline{3-8}
     &  & Prec. & Rec.  & F1 & Prec. & Rec.  & F1  \\
     \midrule
     \multirow{9}{*}{SIGHAN15} & BERT~\cite{bert}       &74.2 &78.0 &76.1 &71.6 &75.3 &73.4  \\
                               & SpellGCN~\cite{spellgcn} &74.8 &80.7&77.7  & 72.1 & 77.7 & 75.9  \\
                               & REALISE~\cite{realise}  &77.3 &81.3  &79.3          & 75.9 & 79.9 & 77.8 \\
                               & MDCSpell~\cite{mdcspell} &\textbf{80.8} &80.6 &80.7  &\textbf{ 78.4} & 78.2 & 78.3 \\
                               & LEAD~\cite{lead}  &\underline{79.2} &82.8 &80.9 & 77.6 & 81.2 & 79.3 \\
                               & ReLM~\cite{relm}  &- &- &- & 73.0 & 81.2 & 76.9 \\
                               & DORM~\cite{liang-etal-2023-disentangled} & 77.9 &\underline{84.3} &\underline{81.0} & 76.6 & \underline{82.8} & \underline{79.6} \\
                           & SCOPE~\cite{scope}* & 78.3 &82.6 &80.4 & 76.5 & 80.8 & 78.6 \\
     \cline{2-8}
                               & COIN (BERT)             & -& -& -    & 72.7 & 81.9 & 77.0     \\
                               & COIN (SCOPE)           &\underline{79.2} &\textbf{84.5} &\textbf{81.8}      & \underline{78.2} & \textbf{83.4} & \textbf{80.7}      \\

    \bottomrule
    
    \end{tabular}
    }
    \label{table4}
    \vspace{0mm}
    
\end{table*}

\subsection{Baseline methods}

We compare our method with the following baselines:

\noindent \textbf{BERT} ~\cite{bert} uses the word embedding as the softmax layer on the top
of BERT for the CSC task. We use the same masking strategy on non-error tokens as ~\cite{rethinking} for better performance. 

\noindent \textbf{REALISE} ~\cite{realise} selectively mixes the semantic, phonetic and graphic information of Chinese characters as the input of a correct network.

\noindent \textbf{SpellGCN} ~\cite{spellgcn} employs GCN to incorporate phonetic and visual knowledge and model the character similarity for CSC task.

\noindent \textbf{MDCSpell} ~\cite{mdcspell} designs a multi-task framework, where BERT is used as a corrector to capture the visual and phonological features of the characters and integrated the hidden states of the detector to reduce the impact of error. We use the same masking strategy on non-error tokens as ~\cite{rethinking} for better performance. 

\noindent \textbf{LEAD} ~\cite{lead} models phonetic, visual, and semantic information by a contrastive learning framework.

\noindent \textbf{DORM} ~\cite{liang-etal-2023-disentangled} designs a self-distillation module which disentangle two types of feature for direct interaction and predict character from pinyin.

\noindent \textbf{ECSpell} ~\cite{ecspell} uses an error consistency masking strategy which is used to specify the error types of automatically generated sentences, and attachs a User Dictionary guided inference module to a general token classification based speller.

\noindent \textbf{RSpell} ~\cite{rspell} employs pinyin fuzzy matching to search terms to create combination inputs and introduces an adaptive process control mechanism to dynamically adjust the impact of external knowledge on the model. 

\noindent \textbf{ReLM} ~\cite{relm} trains model to rephrase the entire sentence by infilling additional slots, instead of character-to-character tagging, to imitate human mindset.

\subsection{Training Details}

We load the weight of Chinese BERT. Following ~\cite{rethinking}, we also use the confusion set in ~\cite{plome} to synthesize paired sentences in \textit{wiki2019zh} and \textit{news2016zh} for pre-training. 

Otherwise, our method does not focus on the application of phonetic or visual information of characters, which brings certain improvement in some previous works. However, our method can be easily attached to existing methods by only changing the backbone. So we replace the BERT in our vanilla method by SCOPE~\cite{scope}, which takes into account phonetic and visual information. Then we name it COIN(SCOPE) to distinguish it from COIN(BERT).


We empirically set the thresholds to achieve precision and recall of 0.95 separately in high precision results and in high recall results. This choice of 0.95 reflects a common confidence level in statistical analysis. The code is available in GitHub\footnote{\url{https://github.com/GreedyGeorge/COIN}}.

\subsection{Experiments on ECSpell}

 Table \ref{table3} sumarries our results on three domains of ECSpell dataset. We can see that COIN consistently outperforms all the baselines in all metrics, verifying its effectiveness for CSC task. The improvemenets are substantial in these datasets comparing with sota method, $e.g.$, +2.6 in LAW, +7.7 in MED and +5.0 in ODW.

 Additionally, our method achieves both higher precision and recall, demonstrating that it benefits from the synergy of two strategic implementations. The improvements of this dual-strategy approach are not the result of a compromise between precision and recall. Instead, it is a simultaneous enhancement in both metrics, underscoring the effectiveness of integrating precise error position with adaptive context awareness in the error correction process. This configuration optimally leverages the strengths of both strategies, confirming that the enhancements in model performance are additive rather than compensatory.

\subsection{Experiments on SIGHAN}

Table \ref{table4} sumarrizes our results on SIGHAN15 dataset. COIN also gets the best result comparing with others in SIGHAN15. Despite only using BERT as the backbone network, our approach still achieves competitive results. Comparing with backbones, COIN(BERT) improves the F1 score by 3.6 and COIN(SCOPE) improves the F1 score by 2.1, which proves that our method is easy to expand and effective.

In our experiments, we present the raw results of SCOPE without any post-processing. Notably, SCOPE benefits significantly from a well-designed post-processing strategy, which can also be applied to other methods as a standalone enhancement. To ensure fairness in comparison with other methods that do not employ such post-processing, we have omitted this strategy from our experiments.

\section{Analysis and Discussion}

\subsection{Ablation Study}

Our method primarily benefits from the two strategies of using high precision and high recall results. Therefore, we conduct ablation studies on ECSpell LAW dataset with the following settings: (1) removing the selective masking strategy, (2) removing the error position information fusion strategy. For the former, concatenating a same sentence to the source sentence then predicting two concatenated target sentences is meaningless. So we mask the entire sentence that is concatenated to source sentence, $i.e.$, we concatenate the source sentence with a series of [MASK] tokens equal in length to it.

To prove that using Fuzzy Indication strategy can instruct the model to focus on the context of errors and then improve the effect of error correction, we also remove Fuzzy Indication strategy, which means that the error position information is directly used in feature fusion.

\begin{table}[t]
\small
\caption{Results of ablation experiments in ECSpell Law dataset. "\textit{\,w/o} FI" means adding the error position directly without Fuzzy Indication processing. "\textit{\,w/o} EP" means that error position information is not used in feature fusion. "\textit{\,w/o} SM" means that selective masking strategy is not used. In parentheses is the gap compared to the complete method. }
\vspace{2em}
\centering
    \resizebox{0.6\columnwidth}{!}{
        \begin{tabular}{l|c c c}
            \toprule
            \multirow{2}{*}{{Method}}  &  \multicolumn{3}{c}{{ECSpell Law}} \\
            \cline{2-4}
            \\[-0.9em]
                                         & {Prec.}     & {Rec.}        &      {F1}   \\
            \midrule
            \\[-0.9em]
            \textbf{COIN}               & \textbf{93.5}    &   \textbf{96.1}   &   \textbf{94.8}  \\
            \multirow{2}{*}{\ \textit{\, w/o} FI}                          & 93.0   & 93.7  & 93.4 \\
             & (-0.5)  & (-2.4)  &  (-1.4)\\
            \multirow{2}{*}{\ \textit{\, w/o} EP  }                        & 91.6   & 94.9   & 93.2  \\ 
            & (-1.9)  &  (-1.2)  &  (-1.6) \\
            \multirow{2}{*}{\ \textit{\, w/o} SM }              & 92.0 & 94.1  & 93.0 \\
            & (-1.5) & (-2.0)  &  (-1.8) \\
            \hline
            \multirow{2}{*}{\ \textit{\, w/o} EP\&SM }                    & 86.2  & 97.7   & 91.5  \\
             &  (-7.3) & (+1.6)  & (-3.3) \\
           
            \bottomrule
        
        \end{tabular}
    }
    
    \label{table5}
    \vspace{0mm}
    
\end{table}

The results are presented in Table \ref{table5}. Our experiments have demonstrated that the removal of any strategies results in a deterioration of performance. Additionally, the use of Fuzzy Indication proved to be effective. Note that when we omit one of EP and SM, both precision and recall are decrease, which proves that the two strategies function collaboratively rather than merely leveraging high-precision results to enhance error correction precision or relying solely on high-recall results to boost error correction recall. In the "point out the errors and fill in the blanks" game, both the "point out" and "blank" provide critical information about the occurrence of errors. 
This is why using only one of them shows a similar improvement. 

Additionally, when both of these strategies are omitted, the recall metric improves by 1.6. This can be explained as that when every source character is masked and no any indication of error position is given, the model may engage in additional correction, which means that the model will correct more characters even though they are probably right. This leads to a slight improvement in recall but a significant reduction in precision, as evidenced by a decrease of 7.3 in our experiments.

\subsection{Strategy Setting}

For FI, we experiment with different distributions. As shown in Table \ref{table7}, Gaussian Distribution yields the best result. Notice that Uniformly Distribution get higher F1 score than Dirac Distribution, highlighting the effectiveness of indicating error context. Moreover, it is found that Triangular Distribution, which is more similar to Gaussian Distribution, works better than Uniformly Distribution. This further supports that Gaussian Distribution is the optimal choice for Fuzzy Indication.

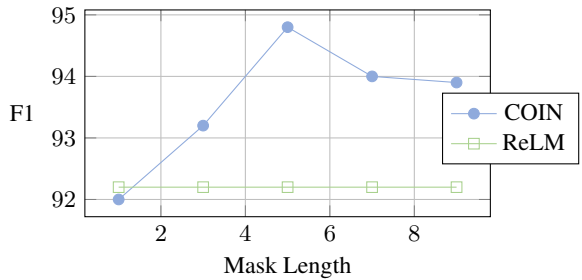
\begin{figure}[t]
    \centering
    \begin{tikzpicture}
        \begin{axis}[
            width=0.8\linewidth,
            height=0.5\linewidth,
            xlabel={Mask Length},
            ylabel={F1},
            ylabel near ticks,
            xlabel near ticks,
            ylabel style={rotate=270, anchor=east},
            grid=both,
            legend style={at={(1.05,0.6)},anchor=north}
        ]
        
        \addplot[mark=*,myblue] coordinates {
            (1,92.0)
            (3,93.2)
            (5,94.8)
            (7,94.0)
            (9,93.9)
        };
        \addlegendentry{COIN}
        
        \addplot[mark=square,mygreen] coordinates {
            (1,92.2)
            (3,92.2)
            (5,92.2)
            (7,92.2)
            (9,92.2)
        };
        \addlegendentry{ReLM}
        
        \end{axis}
    \end{tikzpicture}
    \caption{Different lengths of mask. To ensure symmetry, our experiments are all odd in mask lengths.}
    \vspace{2em}
    \label{fig4}
\end{figure}

For SM, we experimented with different mask lengths. The average sentence length is approximately 30 tokens, $e.g.$, 30.71 in Law, 32.45 in Med, 26.67 in Odw and 30.76 in SIGHAN15. As shown in Figure \ref{fig4}, we experimented with 3 to 9, representing 10\% to 30\% of the average length. The mask length of 5 yields the best result.


\begin{table}[t]
\small
\caption{Different distributions of FI. DD is Dirac Distribution, UD is Uniformly Distribution, TD is Triangular Distribution, and GD is Gaussian Distribution. }
\vspace{1em}
\centering
    \resizebox{0.45\columnwidth}{!}{
        \begin{tabular}{l|c c c}
            \toprule
            \multirow{2}{*}{{FI}}  &  \multicolumn{3}{c}{{ECSpell Law}} \\
            \cline{2-4}
            \\[-0.9em]
                                         & {Prec.}     & {Rec.}        &      {F1}   \\
            \midrule
            \\[-0.9em]
            \textbf{\textit{GD} }               & \textbf{93.5}    &   \textbf{96.1}   &   \textbf{94.8}  \\
            {\textit{TD} }              & 93.1 & 94.9  & 94.0 \\
            {\textit{UD} }                        & 91.7   & 95.7   & 93.7  \\
            {\textit{DD} }                          & 93.0   & 93.7  & 93.4 \\

            \bottomrule
        
        \end{tabular}
    }
    
    \label{table7}
    \vspace{0mm}
    
\end{table}

\begin{CJK*}{UTF8}{gbsn}

\begin{table}[t]\small
\caption{An example of Chinese spelling error. The characters in blue are detected as error in high precision result, that is High P in the table. The characters underlined are detected as error in high recall result, that's High R in the table. The characters in red means the results of correction. }
\vspace{1em}
\centering
\resizebox{0.9\linewidth}{!}{

\begin{tabular}{ll}
\toprule

\makebox[0.15\linewidth][c]{Source}  &\makebox[0.66\linewidth][l]{你可以告诉我\underline{那}家书店有中文\underline{数}\underline{\textcolor{blue}{马}}？} \\
&Can you tell me if there are any Chinese\\& \underline{number} \underline{\textcolor{blue}{horse}} in \underline{that} bookstore? \\
\midrule
\makebox[0.15\linewidth][c]{Target}  &\makebox[0.66\linewidth][l]{你可以告诉我那家书店有中文\textcolor{red}{书}\textcolor{red}{吗}？} \\
& Can you tell me if there are any Chinese\\& \textcolor{red}{books} in that bookstore?\\

\midrule
\makebox[0.15\linewidth][c]{High P} & 0, 0, 0, 0, 0, 0, 0, 0, 0, 0, 0, 0, 0, 0, \textcolor{blue}{1}, 0 \\
\midrule
\makebox[0.15\linewidth][c]{High R} & 0, 0, 0, 0, 0, 0, \underline{1}, 0, 0, 0, 0, 0, 0, \underline{1}, \underline{1}, 0  \\
\midrule
\makebox[0.15\linewidth][c]{Output}  &\makebox[0.66\linewidth][l]{你可以告诉我\underline{那}家书店有中文\underline{\textcolor{red}{书}\textcolor{red}{吗}}？} \\
&Can you tell me if there are any Chinese\\& \underline{\textcolor{red}{books}} in \underline{that} bookstore?\\
\bottomrule
\end{tabular}
}

\label{table6}
\end{table}

\end{CJK*}

\begin{table}[h!]
\small
\caption{Overall results of LLMs and our method on SIGHAN15 in precision, recall, and F1. The best results are shown in bold. }
\vspace{1em}
    \centering
    \resizebox{0.95\linewidth}{!}{
    \begin{tabular}{c | l | p{1cm}<{\centering} p{1.0cm}<{\centering} p{0.9cm}<{\centering} }
    \toprule
    {Dataset} & {Methods} &    Prec. & Rec.  & F1 \\
     \midrule
     \multirow{9}{*}{\makecell{SIGHAN\\15}} & ChatGLM-6B      & 1.4 & 2.0 & 1.6 \\
                               & ChatGLM2-6B            & 2.4 & 3.9 & 3.0  \\
                               & Vicuna-13B-v1.3  & 2.6 & 3.3 & 2.9 \\
                      
                      & Baichuan-13B-Chat  & 6.1 & 10.4 & 7.7 \\
                               & GPT-3.5-Turbo    & 14.3 & 25.1 & 18.2 \\
                           & text-davinci-003  & 15.4 & 26.6 & 19.5 \\
                           & ERNIE Bot  & 38.7 & 34.2 & 36.4 \\
                           & GPT-4  & 44.4 & 36.3 & 40.3 \\
     \cline{2-5}
                               & COIN (ours)                 & \textbf{78.2} & \textbf{83.4} & \textbf{80.7}   \\

    \bottomrule
    
    \end{tabular}
    }
    \label{table7}
    \vspace{0mm}
    
\end{table}

\subsection{Case Study}
\begin{CJK*}{UTF8}{gbsn}

As is shown in Table \ref{table6}, given a Chinese sentence with two spelling errors ``数" and ``马", ``马" is included but ``数" is missing in high precision result, while both of them are included as well as a right character ``那" in high recall result. By using the two strategies of our method, ``马" is masked and corrected to be ``吗" when it is also been indicated as error in information fusion, ``数" is masked and corrected to be ``书" although it is not been indicated as error in information fusion, and ``那" is preserved although it is masked. We can see that although a wrong character with high uncertainty is not been detected in high precision result, it is probably to be detected in high recall result, which also provides the model with indications of error. In other words, most characters in high precision result is wrong and most characters not in high recall result is right. So the model can easily judge to keep or replace a character by the assistant of two detection results.

\begin{figure}[t]
\centering
\includegraphics[width=1\linewidth]{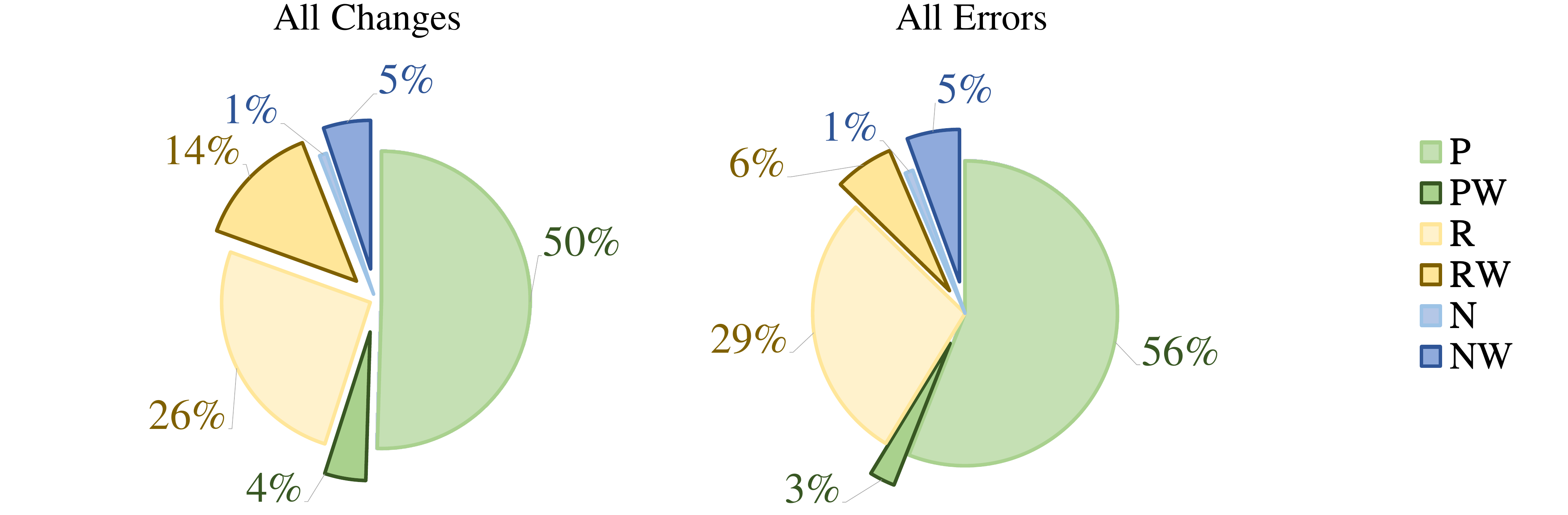}
\caption{Proportion of correct corrections among various error detection results in SIGHAN15. "P" signifies the errors in high precision results, "R" signifies in high recall results, "N" signifies not been detected. The suffix "W" indicates cases where correction has failed.}
\vspace{2em}
\label{fig5}
\end{figure}

We explore this phenomenon and show it in Figure \ref{fig5}. We performed a statistical analysis of all changed characters and all erroneous characters in source sentences. In both all changed characters and all existing errors, most of errors in high precision detection results are corrected as expected, and most of errors not in high precision results but in high recall results are also corrected.

\subsection{Experiments on LLMs}
Considering the amazing ability of large language models (LLMs) on various NLP tasks, we test the ability of LLMs on CSC task in this section. Following the experimental setups of \cite{li2023ineffectiveness}, we compare the performance of some existing LLMs. The results are shown in the Table \ref{table7}. 

It is shown that the best performing LLM, GPT-4, still has a large gap with our method on the CSC task. In our experiments, we find that even with well-designed prompt, LLMs may output sentences that are logically correct but do not meet the existing indicators of CSC task. For example, as shown in Table \ref{table8}, GPT-4 outputs a sentence with the similar meaning as the answer and without spelling errors, but it adds extra words rather than only correct the spelling error. Also, GPT-4 does not take into account the causes of typo generation, such that ``跑" has the similar phoneme and glyph with ``饱", instead it rewrite the sentence through its own understanding, which may be a reason of the poor performance.
\end{CJK*}

\begin{CJK*}{UTF8}{gbsn}

\begin{table}[t]\small
\caption{An example of corrections that does not meet the existing metrics in SIGHAN15. The characters in red mean wrong and in blue means right. }
\vspace{1em}
\centering
\resizebox{0.95\linewidth}{!}{

\begin{tabular}{ll}
\toprule

\makebox[0.15\linewidth][c]{Source}  &\makebox[0.66\linewidth][l]{他睡很\textcolor{red}{跑}，睡到忘了时间起床。} \\
&He slept so \textcolor{red}{run} that he forgot the time to get up. \\
\midrule
\makebox[0.15\linewidth][c]{Target}  &\makebox[0.66\linewidth][l]{他睡很\textcolor{blue}{饱}，睡到忘了时间起床。} \\
& He slept so \textcolor{blue}{much} that he forgot the time to get up.\\

\midrule

\makebox[0.15\linewidth][c]{GPT-4}  &\makebox[0.66\linewidth][l]{他睡\textcolor{red}{得}很\textcolor{red}{晚}，睡到忘了时间起床。} \\
&He slept so \textcolor{red}{late} that he forgot the time to get up.\\
\bottomrule
\end{tabular}
}

\label{table8}
\end{table}

\end{CJK*}

\vspace{0mm}
\section{Conclusion}
\vspace{0mm}

This work proposes a Chinese Spelling Correction method based on a detector-corrector framework. Recognizing the limitations of detector that simultaneously improving precision and recall is impossible, we constrain the detector for desired results and design two strategies which are combined in the correction. Furthermore, we consider the contextual relevance of error positions and provide explicit guidance in the application of detection results to the correction process, also enabling our error indication method to handle errors with uncertain positions within Chinese phrases. Extensive experiments confirm the effectiveness of our method.

\newpage



\bibliography{mybibfile}

\end{document}